\documentclass[letterpaper, 10 pt, conference]{ieeeconf}  

\IEEEoverridecommandlockouts                              

\usepackage{amsmath} 

\usepackage[hidelinks]{hyperref}
\usepackage[capitalise]{cleveref}

\usepackage{tikz}
\usepackage{pgfplots}
\usepackage{pgfplotstable}
\usepackage{ctable}
\usetikzlibrary{external}
\tikzexternaldisable

\usepackage{xparse}

\usepackage{multirow}

\title{\LARGE \bf
Analyzing the Cross-Sensor Portability of Neural Network Architectures for LiDAR-based Semantic Labeling
}

\author{Florian Piewak$^{1,2}$, Peter Pinggera$^{1}$, and Marius Z\"ollner$^{2,3}$
	\thanks{$^{1}$Daimler AG, R\&D, Stuttgart, Germany}%
	\thanks{$^{2}$Karlsruhe Institute of Technology, Karlsruhe, Germany}%
	\thanks{$^{3}$FZI Research Center for Information Technology, Karlsruhe, Germany}%
	\thanks{Primary contact: {\tt\small florian.piewak@daimler.com}}%
}

\begin{document}

\maketitle
\thispagestyle{empty}
\pagestyle{empty}

\newcommand{\equals}{\mathrel{\widehat{=}}}

\newcommand\pointl{P}
\newcommand\xGlobal{x_{g}}
\newcommand\yGlobal{y_{g}}
\newcommand\zGlobal{z_{g}}
\newcommand\reflectivity{r}
\newcommand\xVoxel{x_{v}}
\newcommand\yVoxel{y_{v}}
\newcommand\zVoxel{z_{v}}

\newcommand\numXVoxel{num_{x}}
\newcommand\numYVoxel{num_{y}}
\newcommand\numZVoxel{num_{z}}

\newcommand\lilanet{\textit{LiLaNet}}
\newcommand\pillarlikenet{\textit{PiLaNet}}
\newcommand\autolabeling{\textit{Autolabeling}}

\definecolor{road}{RGB}{128, 64, 128}
\definecolor{sidewalk}{RGB}{244, 35, 232}
\definecolor{person}{RGB}{220, 20, 60}
\definecolor{rider}{RGB}{255, 0, 0}
\definecolor{smallVehicle}{RGB}{0, 0, 142}
\definecolor{largeVehicle}{RGB}{0, 0, 70}
\definecolor{twoWheeler}{RGB}{119, 11, 32}
\definecolor{construction}{RGB}{70, 70, 70}
\definecolor{pole}{RGB}{153, 153, 153}
\definecolor{trafficSign}{RGB}{220, 220, 0}
\definecolor{vegetation}{RGB}{107, 142, 35}
\definecolor{terrain}{RGB}{152, 251, 152}
\definecolor{sky}{RGB}{70, 130, 180}

\begin{abstract}
State-of-the-art approaches for the semantic labeling of LiDAR point clouds
heavily rely on the use of deep Convolutional Neural Networks (CNNs).
However, transferring network architectures across different LiDAR
sensor types represents a significant challenge,
especially due to sensor specific design choices
with regard to network architecture as well as data representation.
In this paper we propose a new CNN architecture
for the point-wise semantic labeling of LiDAR data which
achieves state-of-the-art results while increasing portability across sensor types.
This represents a significant advantage given
the fast-paced development of LiDAR hardware technology.
We perform a thorough quantitative cross-sensor analysis
of semantic labeling performance in comparison to a state-of-the-art
reference method.
Our evaluation shows that the proposed architecture is indeed highly portable,
yielding an improvement of 10 percentage points in the Intersection-over-Union (IoU) score
when compared to the reference approach.
Further, the results indicate that the proposed network architecture
can provide an efficient way for the automated generation of large-scale training data
for novel LiDAR sensor types without the need for 
extensive manual annotation or multi-modal label transfer.
\end{abstract}

\section{INTRODUCTION}
\label{sec:introduction}

Within the field of autonomous driving, vehicles
employ multiple sensor modalities such as cameras, RaDAR,
and LiDAR to reliably perceive the environment even in challenging scenarios
and adverse environmental conditions \cite{Aeberhard2015,Ziegler2014,Urmson2009,Thrun2006}.
Each sensor modality has to exploit its specific strengths
in order to contribute to the overall geometric and semantic understanding of the scene.
The extracted information is then combined into a holistic 
representation of the environment, for example an occupancy grid map \cite{Nuss2016}.
This environmental representation provides the basis for high-level tasks such as
localization \cite{Vu2011},
situation analysis \cite{Laugier2011} or path planning \cite{Bai2015}.

\begin{figure}
	\centering
	\vspace{2mm}
	\includegraphics[width=\linewidth]{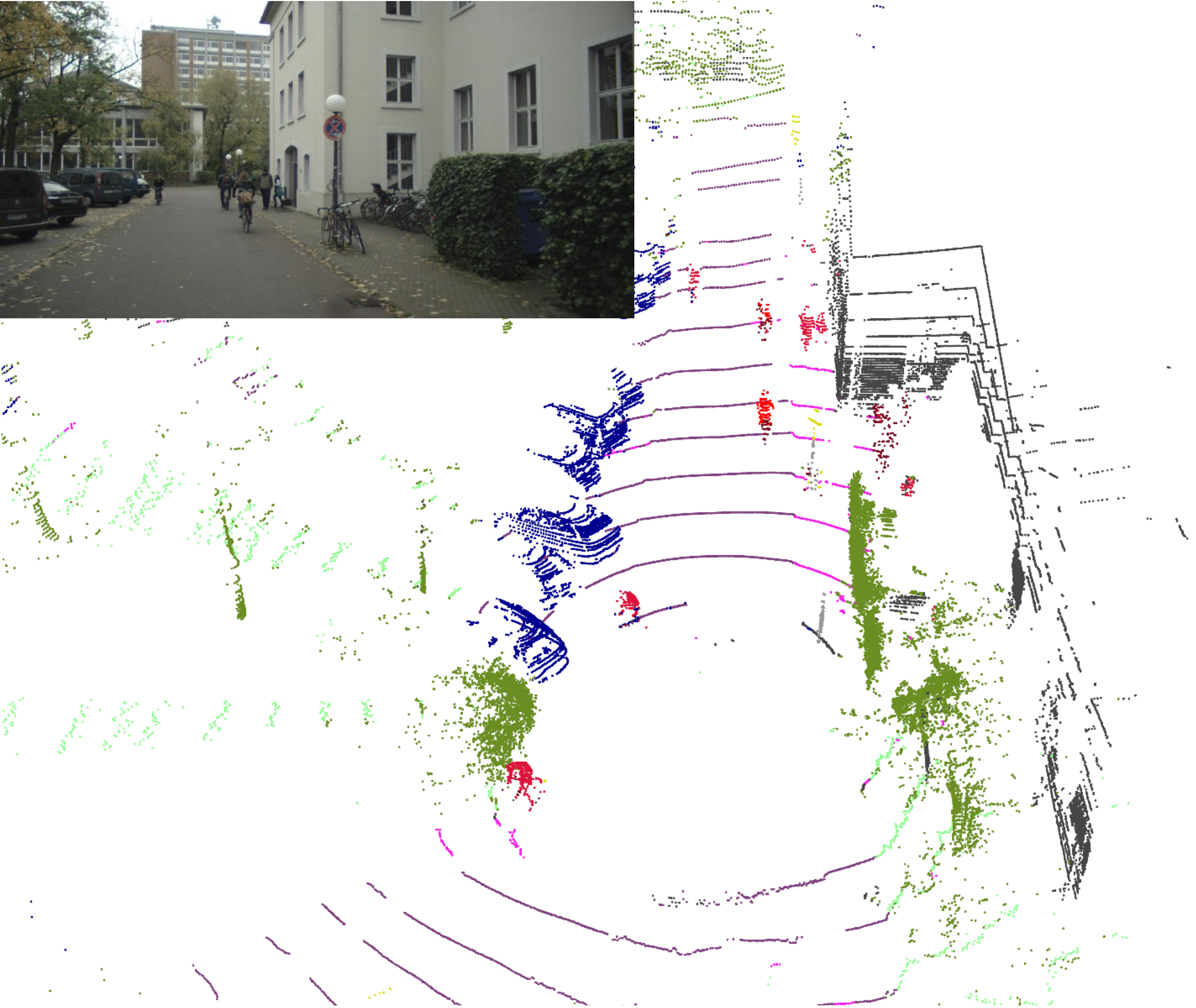}
	\caption{Exemplary semantic labeling result obtained via the proposed \pillarlikenet{} on a VLP-32C point cloud. The corresponding camera image on the top left is shown for clarity, with the camera's field of view covering the top center of the point cloud. The following semantic classes are visualized:
		\textcolor{road}{road},
		\textcolor{sidewalk}{sidewalk},
		\textcolor{person}{person},
		\textcolor{rider}{rider},
		\textcolor{smallVehicle}{small vehicle},
		\textcolor{largeVehicle}{large vehicle},
		\textcolor{twoWheeler}{two wheeler},
		\textcolor{construction}{construction},
		\textcolor{pole}{pole},
		\textcolor{trafficSign}{traffic sign},
		\textcolor{vegetation}{vegetation},
		\textcolor{terrain}{terrain}.}
	\label{fig:pointcloudvlp32}
\end{figure}

\begin{figure}
	\centering
	\vspace{2mm}
	\includegraphics[width=\linewidth]{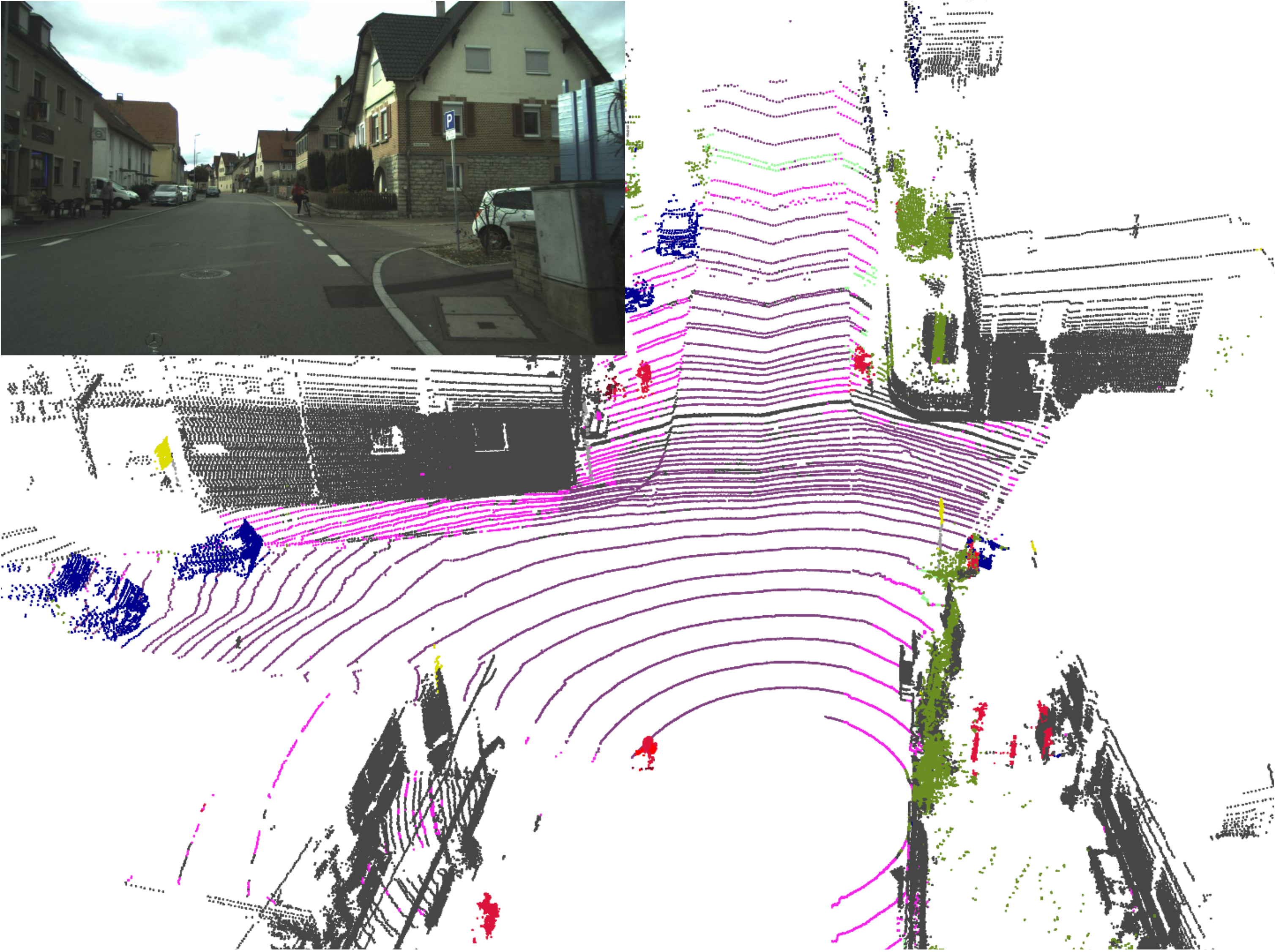}
	\caption{Exemplary semantic labeling result obtained via the proposed \pillarlikenet{} on a VLS-128 point cloud. The corresponding camera image on the top left is shown for clarity, with the camera's field of view covering the top center of the point cloud. The following semantic classes are visualized:
		\textcolor{road}{road},
		\textcolor{sidewalk}{sidewalk},
		\textcolor{person}{person},
		\textcolor{rider}{rider},
		\textcolor{smallVehicle}{small vehicle},
		\textcolor{largeVehicle}{large vehicle},
		\textcolor{twoWheeler}{two wheeler},
		\textcolor{construction}{construction},
		\textcolor{pole}{pole},
		\textcolor{trafficSign}{traffic sign},
		\textcolor{vegetation}{vegetation},
		\textcolor{terrain}{terrain}.}
	\label{fig:pointcloudvls128}
\end{figure}

A key capability to obtain an accurate semantic understanding of an observed scene is the
semantic segmentation of the sensor information.
For each sensor measurement a semantic class is inferred,
yielding a fine-grained prediction of the observed obstacle types (see \cref{fig:pointcloudvlp32} and \cref{fig:pointcloudvls128})
The individual predictions can then be combined into a compact object-level representation
\cite{Barea2018}.

The task of semantic segmentation or semantic labeling originated in the field of computer vision,
with the aim to individually classify each pixel
in a given image \cite{Garcia-Garcia2017}. Within recent years, this task has
been applied successfully to other sensor modalities such as LiDAR or RaDAR
\cite{Feng}. Across all modalities state-of-the-art results are obtained by
modern deep learning techniques.
However, approaches based on the application of deep Convolutional Neural Networks (CNNs)
are often tailored to the specific characteristics of the respective sensor instance.
Transferring a network architecture from e.g. one LiDAR
sensor model to another represents a significant challenge,
especially due to sensor specific design choices
with regard to network architecture as well as data representation.
This effect is intensified by the fact that LiDAR sensor technology keeps evolving at a fast pace,
with numerous new sensor types being announced or released to the market every year,
featuring novel scanning patterns and ever increasing range and spatial resolution.

When training deep neural networks, large scale datasets are required
in order to obtain state-of-the-art results.
Generating manually annotated point cloud data for LiDAR-based semantic labeling
at scale presents a vast effort and involves even higher cost than
manual image annotation in the computer vision domain. This is due to the additional
spatial dimension as well as the sparsity of the data, which yields a representation
that is non-intuitive and tedious for human annotators.
Therefore, some authors have used existing datasets dedicated to other tasks
such as 3D object detection \cite{Geiger2012} to extract point-wise
LiDAR semantics \cite{Wu2018,Dewan2017}.
However, only a small number of semantic classes can be extracted in this way.
An alternative approach to generate 3D training data
is based on the transfer of semantic
information from a registered camera image to the LiDAR domain
\cite{Varga2017,Piewak2018}. While this allows the automated generation of
large amounts of training data, the quality of the result strongly depends on
the accuracy of the image-based semantics,
the extrinsic and intrinsic calibration as well as the temporal synchronization
of the different sensor modalities.\\

Given the sensor specific tailoring of many current CNN architectures as well as
the difficulty of generating annotated training data at scale,
in this paper we consider the cross-sensor portability of neural network architectures for
LiDAR-based semantic labeling.
Portable network architectures could provide a solution to above challenges by
enabling the reuse of annotated data
and reducing the effort required for adapting CNN architectures to new sensor models.

Our main contributions can be summarized as follows:

\begin{enumerate}
	\item We present a portable Convolutional Neural Network (CNN) architecture
	for high-quality semantic labeling of semi-dense point clouds across different sensor models.
	\item We perform a thorough quantitative cross-sensor evaluation of the
	semantic labeling performance in comparison to state-of-the-art
	approaches.
    Our experimental analysis is based on extensive dedicated datasets from
    two state-of-the-art LiDAR sensors with strongly differing spatial resolution.
\end{enumerate}

\section{RELATED WORK}
\label{sec:related_work}

In recent years LiDAR point cloud processing
has gained more and more attention due to decreasing hardware costs
and an increasing number of sensor makes and models available on the market.
The main difference between the different algorithmic approaches
is the representation of the 3D point cloud to be processed.
Three different types of point cloud representations are commonly found
in the literature:

First, the point cloud can be represented as a
projection of the 3D data to 2D space. For example
Yang et al. \cite{Yang2018} and Beltr\'an et al. \cite{Beltran2018}
use a bird's eye view (BEV) to extract
bounding boxes for different types of objects from the KITTI
dataset \cite{Geiger2012}. Other approaches use a cylindrical
projection of the LiDAR point cloud, e.g. for semantic
segmentation \cite{Dewan2017,Wang2018,Wu2018,Piewak2018,Mei2018}.
In contrast to other projection methods such as the bird's eye view,
the cylindrical projection provides a lossless transformation of the input data.
The main advantage of the described
projection methods is the efficient processing
based on 2D convolutions. However, the generated CNNs are usually
rather sensor specific and can only be transferred to different sensor models
with significant effort.

An alternative method to represent point clouds is by way of an unordered point set \cite{Qi2016,Qi2017,Li2018}. Qi et al.
\cite{Qi2016} propose the so-called PointNet, which first processes each point
independently and then extracts global features using
a max pooling operator. Based on that global
feature a classification of the point cloud can be obtained.
While this approach is able to handle an arbitrary number of unordered points,
the resulting performance tend to be problematic
in large-scale outdoor scenarios as are usually encountered in the
field of autonomous driving.

The third common type of point cloud representation
performs a discretization of the 3D space into a voxel grid \cite{Maturana2015} or an
octree \cite{Riegler2016}. Rieger et al. \cite{Riegler2016}
evaluate a semantic segmentation task by predicting a single semantic class
for all points within an octree cell. Occupancy information is used as a feature for each voxel.
Zhou et al. propose VoxelNet \cite{Zhou2017} for the task of
object detection. Here the idea of PointNet
\cite{Qi2016} is adopted to extract features from an arbitrary number of points per voxel.
While this representation increases cross-sensor portability due to the regularity of the voxel grid,
the additional dimension and the employed 3D convolutions
drastically increase the training and inference times for larger CNN architectures.
Lang et al. alleviate this problem by introducing PointPillars \cite{Lang2018}.
Here the 3D space is compressed to 2.5D by reducing the
number of voxels along the vertical axis to one, leading to an intermediate structure
resembling pillars.
Similar to BEV-based approaches, this 2.5D pseudo-image can then be processed using 2D convolutions, thereby benefiting from the learned features per pillar.
Both VoxelNet \cite{Zhou2017} and PointPillars \cite{Lang2018} are
optimized for object detection tasks. In this paper we build upon these approaches
and propose a CNN architecture for point-wise semantic segmentation
which is portable across different LiDAR sensors.

\section{METHOD}
\label{sec:method}

\begin{figure*}
	\centering
	\includegraphics[width=\linewidth]{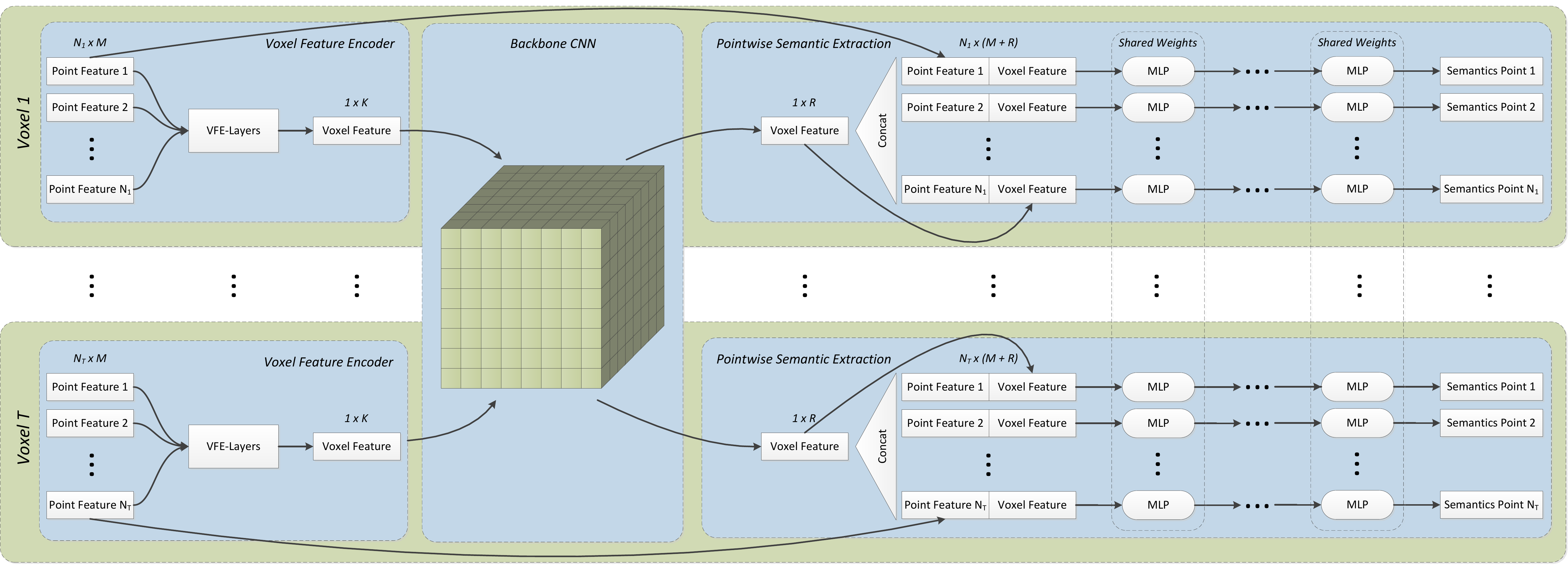}
	\caption{3D voxel CNN architecture for point-wise semantic segmentation. The voxel space as well the processing chain per voxel are represented in green. The three main components (voxel feature encoder, backbone CNN, and point-wise semantic extraction head) are represented as blue boxes.}
	\label{fig:overallCNNarchitecure}
\end{figure*}

\begin{figure*}
	\centering
	\includegraphics[width=\linewidth]{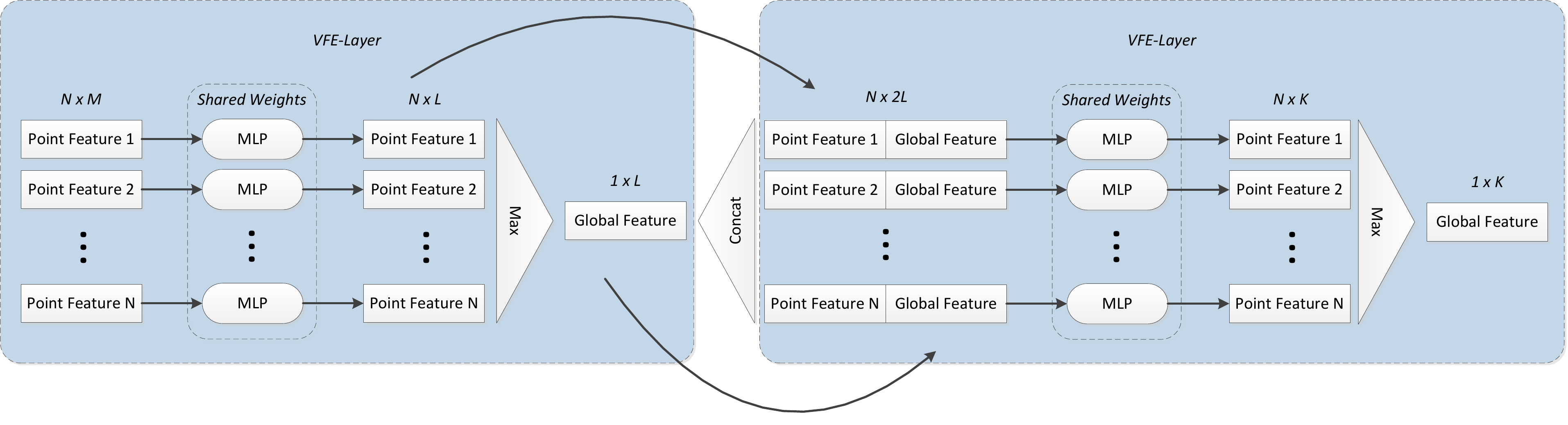}
	\caption{Example of a voxel feature encoder based on two consecutive VFE-Layers.
		The original point features serve as input to the first VFE-Layer, while the second VFE-Layer uses the combination of the point-wise features with the global features as input.}
	\label{fig:multipleVoxelFeatureEncoder}
\end{figure*}

In this paper, we extend previous work and propose a new CNN architecture
for the point-wise semantic labeling of LiDAR data.
The presented architecture achieves state-of-the-art results
while increasing portability across sensor types at the same time.
We choose a voxel-based processing approach, similar to
the data structures used by VoxelNet \cite{Zhou2017} and PointPillars \cite{Lang2018}.
Hence, the resulting pillar-based labeling network is called \pillarlikenet{}.
Within the following subsections, the network architecture
is described in detail. It consists of three main components
(see \cref{fig:overallCNNarchitecure}):
\begin{itemize}
    \item the Voxel Feature Encoder (VFE), which generates a
    feature vector that encodes the properties of the voxel content,
    \item the backbone CNN, which accumulates the generated voxel features in 3D space and
    \item the point-wise semantic extraction head, which infers a semantic label for each
    point from the encoded voxel features.
\end{itemize}

\subsection{Voxel Feature Encoder}
\label{sec:voxelFeatureExtractor}

The voxel feature encoder\footnote{also called voxel feature extractor}
represents a network component designed
to condense the essential properties of all points contained within a voxel
into one feature vector (see \cref{fig:multipleVoxelFeatureEncoder}).
This idea was originally proposed in PointNet \cite{Qi2016} to
encode a full point cloud into a single feature vector.
It was later adapted and applied to individual voxels in the VoxelNet architecture \cite{Zhou2017}.
Each point is represented by its global cartesian coordinates $\xGlobal$,
$\yGlobal$ and $\zGlobal$, the measured reflectivity $\reflectivity$
as well as the relative cartesian coordinates with respect to the mean of all points within a voxel
$\xVoxel$, $\yVoxel$ and $\zVoxel$.
To obtain an initial point-wise feature encoding, each point is processed individually
via multi layer perceptrons (MLP).
These sub-networks use shared weights to enforce identical feature
encoders for each point.
Subsequently a max pooling operator is
applied to the point-wise features in order to generate a single feature
vector per voxel.
The combination of these processing steps is also known as
VFE-Layer in the literature \cite{Zhou2017}.

The VFE-Layer can be applied repeatedly
by concatenating the encoded point-wise features
with the voxel feature vector resulting from the previous step
(see \cref{fig:multipleVoxelFeatureEncoder}).
Eventually the MLPs as well as the max pooling
operator are applied once more to obtained a final refined feature
vector for each voxel.

Since the voxel feature encoder can handle arbitrary numbers
of input points, it is applicable to various voxel sizes
and point cloud densities.
It provides a parameterizable representation
which is highly portable between different sensor types.

The features computed by the VFE-Layers are combined into a 3D
voxel grid which forms the input to the backbone CNN described in the next subsection.

\subsection{Backbone CNN}
\label{sec:backboneCNN}

\begin{figure}
	\centering
	\vspace{2mm}
	\includegraphics[width=\linewidth]{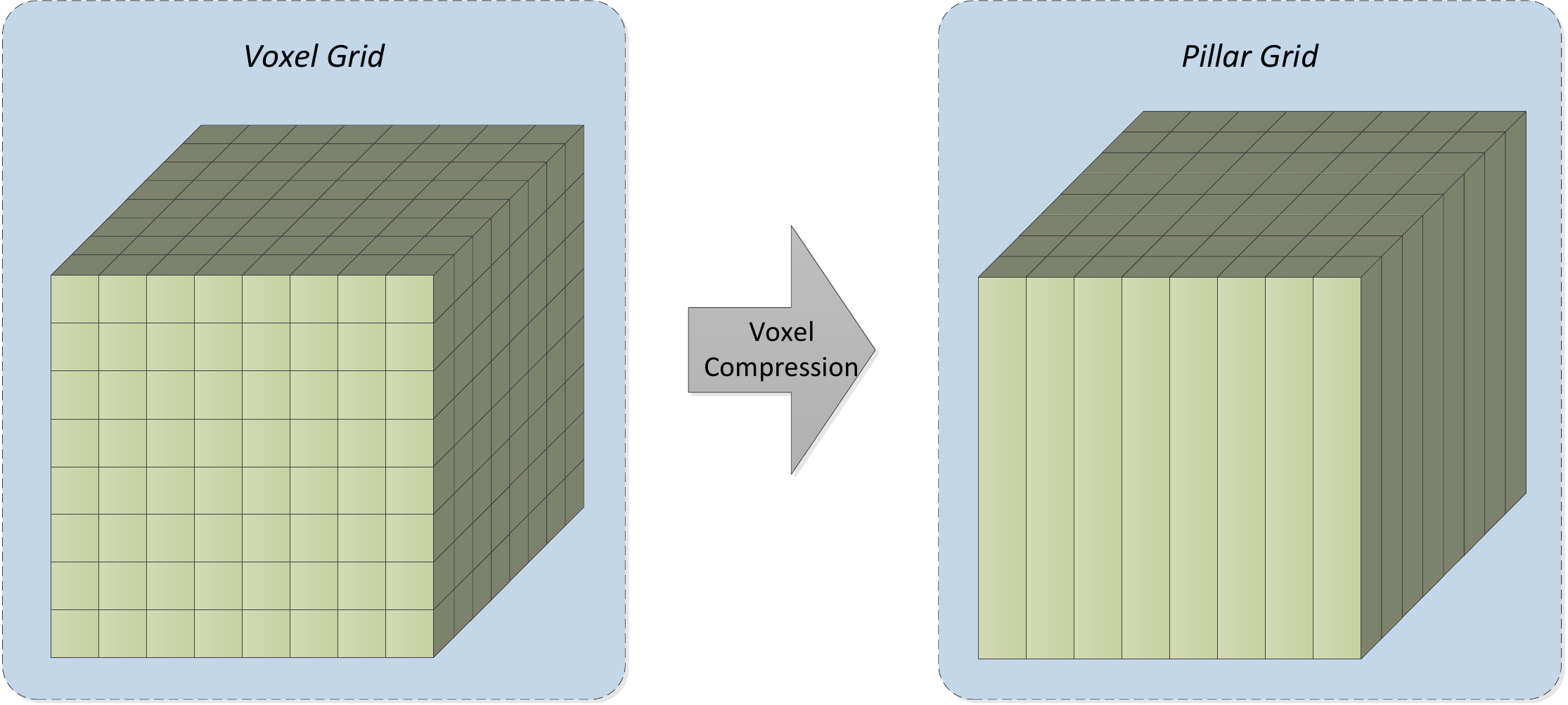}
	\caption{Compression of the 3D space to 2.5D by reducing the
		number of voxels along the vertical axis to one, leading to an intermediate structure resembling pillars as proposed in \cite{Lang2018}.}
	\label{fig:pillars}
\end{figure}

To fully exploit the spatial context within the encoded data,
a CNN backbone architecture is applied to the voxel grid.
Here, various architectural choices are possible.
VoxelNet \cite{Zhou2017} employs a full 3D backbone CNN to extract features
from the voxel representation via 3D convolutions.
This architecture is well suited
for the task of object detection, where
the size of the 3D representation can be reduced
by applying pooling layers or strided convolutions
without a significant loss in output accuracy.
However, for the semantic segmentation task considered in this paper
a fine-grained point-wise prediction is required.
In this case a full 3D CNN entails high computational complexity leading to
excessive memory consumption as well as long training and inference times.
To reduce the dimensionality of the voxel grid and hence the computational complexity
of the backbone CNN, we adopt the concept of pillars
as proposed in \cite{Lang2018} (see \cref{fig:pillars}).
The dimensionality of the voxel grid is reduced by 
directly encoding the height information of each point within its associated pillar.
The resulting representation resembles a two-dimensional pseudo-image,
similar to a bird's eye view.
Consequently, a 2D CNN architecture can be applied
to process the voxel representation in an efficient way while retaining the
full information of the encoded data.

\subsection{Point-Wise Semantic Extraction}
\label{sec:pointwiseSemanticExtraction}

The backbone CNN yields voxel-wise output features,
similar to the probability score maps of object detection approaches
\cite{Zhou2017,Lang2018}.
These features can be used to infer a semantic class for each voxel.
However, to obtain a point-wise semantic segmentation result this approach is not sufficient,
especially for large voxels such as pillars where all
points within a voxel would be assigned the same semantic class.
Therefore, we introduce a point-wise semantic extraction head
as shown in \cref{fig:overallCNNarchitecure}.
For each point, the corresponding output vector
$(\xGlobal, \yGlobal, \zGlobal, \reflectivity, \xVoxel, \yVoxel, \zVoxel)$
of the voxel feature encoder is concatenated with the voxel-wise features
extracted by the backbone CNN.
Subsequently, each point is processed
independently by multiple MLPs to extract a point-wise classification result
independent of its containing voxel.
In this way fine-grained semantic class predictions within one voxel are achieved.

\section{DATASET}
\label{sec:dataset}

In the following analysis of the cross-sensor portability of different network architectures,
we employ extensive dedicated datasets
based on two state-of-the-art LiDAR sensors:
a Velodyne VLP-32C and a Velodyne VLS-128.
\begin{figure}
	\centering
	\vspace{2mm}
	\includegraphics[width=\linewidth]{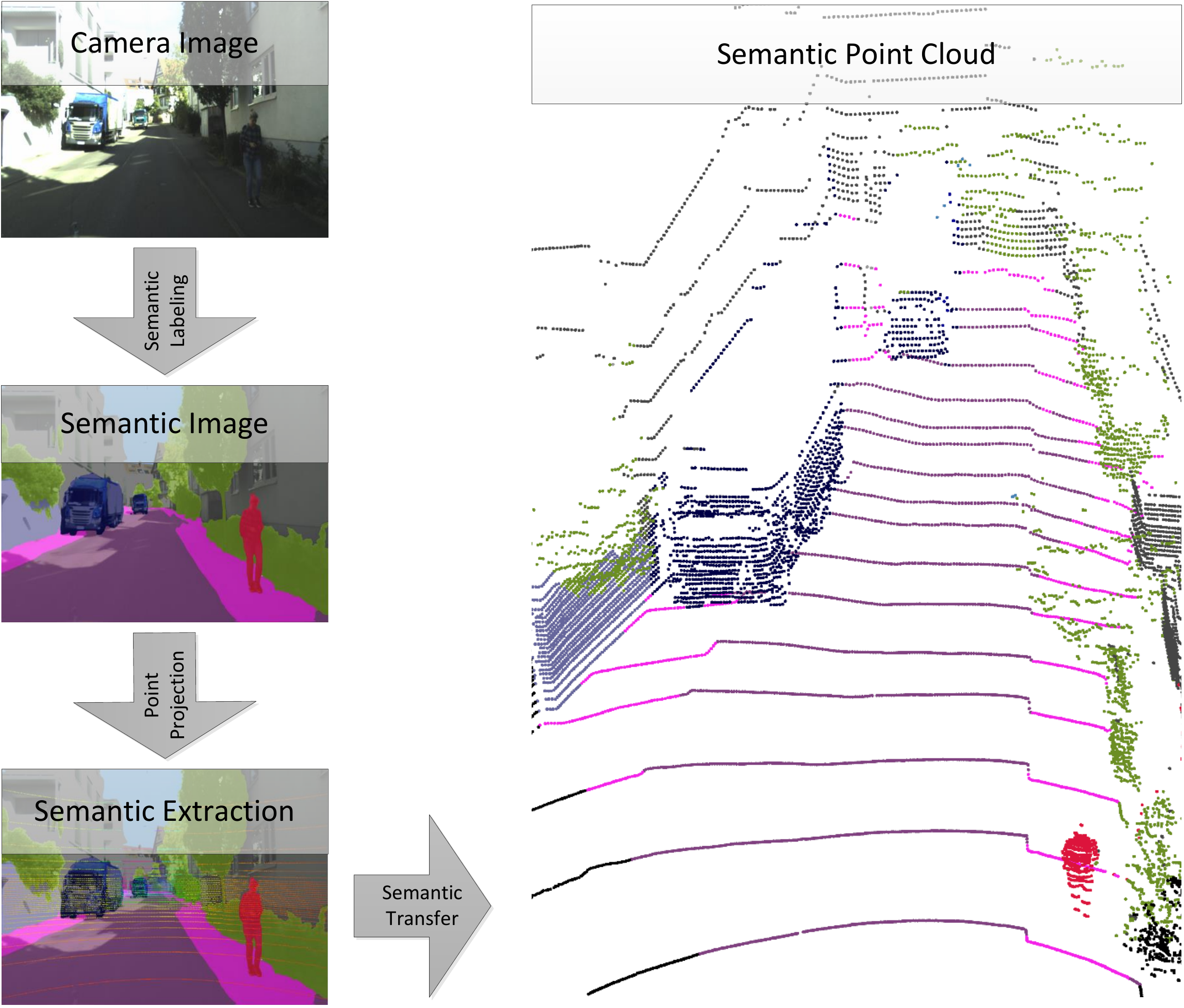}
	\caption{Illustration of the \autolabeling{} method of \cite{Piewak2018} used to transfer semantic reference data from a camera image to a point cloud (VLP-32C).}
	\label{fig:autolabeling}
\end{figure}
The VLP-32C dataset consists of
point clouds with 32 layers (rows)
and is the same dataset as used in \cite{Piewak2018} (see \cref{tab:datasplit}).
The ground truth data is generated in two different
ways: First, the full dataset is semantically annotated in an automated
way using the \autolabeling{} method proposed in
\cite{Piewak2018} (see \cref{fig:autolabeling}).
Here, a dedicated CNN is used to process a registered RGB camera image.
This CNN is trained on the
Cityscapes \cite{Cordts2016} dataset to obtain a high-quality
semantic segmentation result.
The LiDAR point-cloud is then projected
into the camera image and the semantic labels of the image
are transfered to the individual points.
Second, a small number of point clouds is manually annotated
in order to both fine-tune and evaluate the analyzed CNN architectures.
The overall dataset is split into three subsets.
One subset for training the networks (\textit{training}),
one subset for validation as well as optimization of the hyper-parameters of the
architectures (\textit{validation}),
and one subset for the final evaluation (\textit{testing}).

The dataset based on the Velodyne VLS-128, a LiDAR with 128 layers,
is generated in a similar way.
It includes various scenarios such as highways,
rural roads, and urban traffic. The data is also split into three
subsets for training, optimizing hyper-parameters and the final evaluation.
Similar to the VLP-32C dataset, the ground truth is generated in two ways.
The full dataset is annotated automatically via the \textit{Autolabeling}
method, while a small number of frames is manually annotated.
We note that the VLS-128 dataset is approximately half the size of
the VLP-32C dataset (see \cref{tab:datasplit}).

For both datasets the set of semantic classes
is identical to the one defined in \cite{Piewak2018}.
It consists of the following twelve classes:
road, sidewalk, person, rider, small vehicle, large vehicle, two-wheeler,
construction, pole, traffic sign, vegetation, and terrain (see \cref{fig:pointcloudvlp32} and \cref{fig:pointcloudvls128}).

\begin{table}
	\centering
	\vspace{2mm}
	\caption{Split of the two employed datasets into the subsets for training, validation and testing. We distinguish between automatically and manually annotated point clouds (frames).}
	\label{tab:datasplit}
	\begin{tabular}{|c|c|c|c|}
		\hline
		& Training  & Validation & Testing   \\
		\hline
		Autolabeled Frames VLP-32C & $344{,}027$ & $73{,}487$   & $137{,}682$ \\
		\hline
		Autolabeled Frames VLS-128 & $179{,}042$  & $31{,}139$   & $84{,}839$  \\
		\hline
		Man. Annotated  Frames VLP-32C & $1{,}909$   & $373$      & $718$     \\
		\hline
		Man. Annotated  Frames VLS-128 & $1{,}257$   & $335$      & $1{,}110$     \\
		\hline
	\end{tabular}
\end{table}

\section{EXPERIMENTS}
\label{sec:experiments}

In this chapter the proposed network architecture \pillarlikenet{}
is evaluated against a state-of-the-art reference method,
with a focus on cross-sensor portability.
We use the 2D CNN architecture called \lilanet{} of Piewak et al. \cite{Piewak2018}
as our reference.

\pillarlikenet{} implements the voxel representation as described
in \cref{sec:method}, where we limit the voxel space to the range
$(0.0m, -30.0m, -2.0m) \le (\xGlobal, \yGlobal, \zGlobal) \le (60.0m, 30.0m, 9.2m)$.
The number of voxels is set to
$(\numXVoxel, \numYVoxel, \numZVoxel) = (300, 300, 1)$.
These parameters were optimized based
on available GPU memory, overall network performance,
and training time.
The used voxel feature encoder is composed of two VFE-Layers
as described in \cref{sec:voxelFeatureExtractor}, with a voxel
feature vector size of 128. For the implementation of the VFE,
we restrict the maximum number of points per voxel to 35 and
apply random sampling in case the limit is exceeded.
Since we use a single voxel along the z
axis (pillars), the backbone is modeled as a 2D CNN.
For a valid comparison to the reference approach,
we use the same architecture for the backbone CNN as used for \lilanet{} \cite{Piewak2018},
with the number of output features per voxel set to 24.
The point-wise semantic extraction head
includes three consecutive MLP layers with 64,
64, and 12 features. The last layer provides the
scores of the 12 semantic classes.
Aside from the classification score,
after each layer a rectified linear unit (ReLU) is applied.

Note that we do not optimize either of the considered networks architectures
for the specific sensor types.
In this way we achieve a valid evaluation of the unmodified CNN architectures and
corresponding point-cloud representations in terms of cross-sensor portability.\\

The training of both \lilanet{} and \pillarlikenet{}
is performed with a batch size\footnote{In case the network does not fit into GPU memory, the batch is distributed over multiple devices.}
$b = 8$ via the Adam solver \cite{Kingma2014}.
With regard to training strategy, we follow the suggestion of \cite{Piewak2018}
and first train on the autolabeled training subset, followed by
a fine-tuning step on the manually annotated training subset.
The training on the autolabeled set is
run for $300{,}000$ iterations before starting the fine-tuning.
As training parameters we use the suggested default values for
the Adam solver of $\beta_1 = 0.9$, $\beta_2 = 0.999$ and $\epsilon = 10^{-8}$.
The learning rate is fixed at $\alpha = 10^{-3}$ ($\alpha = 10^{-4}$
for fine-tuning), while the weights are initialized with MSRA \cite{He2015}. 

Evaluation is performed based on the testing subset of the
manually annotated frames, where for each class the Intersection over
Union (IoU) metric as well as the overall mean IoU is calculated.
We restrict the evaluation to the defined voxel range to ensure comparability
between the different point cloud representations,
i.e. the voxel representation used by \pillarlikenet{}
and the cylindrical 2D projection of \lilanet{}.

Several evaluation stages are performed on both datasets,
which we discuss in more detail in the following subsections:
\begin{enumerate}
	\item Quality of the \autolabeling{} result
	\item Performance of the networks (\lilanet{} and \pillarlikenet{})
	trained, fine-tuned, and evaluated on the same sensor
	\item Performance of the networks (\lilanet{} and \pillarlikenet{})
	trained and fine-tuned on one sensor and evaluated on the other sensor
	\item Performance of the networks (\lilanet{} and \pillarlikenet{})
	trained and fine-tuned on one sensor and additionally fine-tuned as well as evaluated
	on the other sensor
\end{enumerate} 

\subsection{Autolabeling Quality}
\label{sec:autolabelingPerformance}

The quality of the point cloud semantics obtained by the \autolabeling{} method is
summarized in the first two rows of \cref{tab:overallResults}\footnote{The \autolabeling{} results differ from the values reported in \cite{Piewak2018} due to the restriction to the voxel range.}.
These values provide a reference which
the CNN can be expected to reach without fine-tuning on manually annotated data.
It has to be noted that the results on the VLP-32C dataset are slightly better
than on the VLS-128 dataset. This is due to differences in the
recording setup, including extrinsic calibration data.
The lower resolution of the VLP-32C makes the process less susceptible
to errors caused by calibration inaccuracies.
Nevertheless, the results lie within a similar range
and allow for an initial assessment of the quality of the datasets.

\begin{table*}
	\centering
	\vspace{2mm}
	\caption{Overview of the results obtained in the different evaluation stages described in
		\cref{sec:experiments}. Each row represents a semantic segmentation
		approach. The corresponding descriptions are given in the
		first column, where the first two rows describe the architecture and the training dataset as
		well the additional fine-tuning dataset (if used). The last row of
		the description denotes the dataset used for
		evaluation.
		The top results of the respective network architectures
		trained with the same strategy are marked in bold.}
	\label{tab:overallResults}
	
	\newcolumntype{S}{>{\centering\arraybackslash} m{0.65cm} }
	
	\begin{tabular}{m{2.33cm}|S|S|S|S|S|S|S|S|S|S|S|S||S|}
		& \rotatebox{90}{road} & \rotatebox{90}{sidewalk} & \rotatebox{90}{person} & \rotatebox{90}{rider} & \rotatebox{90}{small vehicle} & \rotatebox{90}{large vehicle} & \rotatebox{90}{two wheeler} & \rotatebox{90}{construction} & \rotatebox{90}{pole} & \rotatebox{90}{traffic sign} & \rotatebox{90}{vegetation} & \rotatebox{90}{terrain} & \rotatebox{90}{mean IoU} \\
		\hline
		\autolabeling{} \newline evaluation VLP-32C
		& $90.3\%$ & $62.4\%$ & $79.7\%$ & $52.7\%$ & $83.1\%$ & $61.9\%$ & $46.5\%$ & $76.5\%$ & $33.6\%$ & $45.1\%$ & $79.6\%$ & $55.5\%$ & $63.9\%$ \\
		\hline
		\autolabeling{} \newline evaluation VLS-128
		& $88.3\%$ & $58.3\%$ & $64.2\%$ & $44.6\%$ & $82.8\%$ & $56.6\%$ & $39.8\%$ & $83.0\%$ & $26.5\%$ & $36.7\%$ & $83.1\%$ & $55.4\%$ & $60.1\%$ \\
		\specialrule{0.5em}{0em}{0em}
		\lilanet{} VLP-32C \newline evaluation VLP-32C
		& $\mathbf{93.9\%}$ & $\mathbf{73.0\%}$ & $72.2\%$ & $45.3\%$ & $86.3\%$ & $49.3\%$ & $47.2\%$ & $\mathbf{84.1\%}$ & $48.3\%$ & $\mathbf{79.0\%}$ & $\mathbf{83.0\%}$ & $\mathbf{66.6\%}$ & $69.0\%$ \\
		\hline
		\pillarlikenet{} VLP-32C \newline evaluation VLP-32C
		& $93.2\%$ & $69.8\%$ & $\mathbf{81.8\%}$ & $\mathbf{50.2\%}$ & $\mathbf{88.9\%}$ & $\mathbf{67.3\%}$ & $\mathbf{47.5\%}$ & $81.6\%$ & $\mathbf{48.8\%}$ & $77.5\%$ & $79.4\%$ & $64.3\%$ & $\mathbf{70.9\%}$ \\
		\specialrule{0.2em}{0em}{0em}
		\lilanet{} VLS-128 \newline evaluation VLS-128
		& $89.9\%$ & $62.3\%$ & $59.5\%$ & $19.1\%$ & $82.1\%$ & $24.8\%$ & $31.5\%$ & $83.5\%$ & $41.3\%$ & $\mathbf{48.3\%}$ & $\mathbf{85.0\%}$ & $\mathbf{65.7\%}$ & $57.7\%$ \\
		\hline
		\pillarlikenet{} VLS-128 \newline evaluation VLS-128
		& $\mathbf{91.1\%}$ & $\mathbf{63.9\%}$ & $\mathbf{69.1\%}$ & $\mathbf{49.7\%}$ & $\mathbf{88.5\%}$ & $\mathbf{41.9\%}$ & $\mathbf{40.0\%}$ & $\mathbf{85.3\%}$ & $\mathbf{47.4\%}$ & $42.2\%$ & $84.5\%$ & $64.2\%$ & $\mathbf{64.0\%}$ \\
		\specialrule{0.5em}{0em}{0em}
		\lilanet{} VLP-32C \newline evaluation VLS-128
		& $55.6\%$ & $19.9\%$ & $17.5\%$ & $7.5\%$ & $45.1\%$ & $4.6\%$ & $9.5\%$ & $65.7\%$ & $33.4\%$ & $\mathbf{36.5\%}$ & $72.7\%$ & $28.0\%$ & $33.0\%$ \\
		\hline
		\pillarlikenet{} VLP-32C \newline evaluation VLS-128
		& $\mathbf{46.5\%}$ & $\mathbf{20.0\%}$ & $\mathbf{48.5\%}$ & $\mathbf{35.4\%}$ & $\mathbf{81.2\%}$ & $\mathbf{34.5\%}$ & $\mathbf{20.3\%}$ & $\mathbf{73.7\%}$ & $\mathbf{41.3\%}$ & $33.9\%$ & $\mathbf{77.1\%}$ & $\mathbf{30.7\%}$ & $\mathbf{45.2\%}$ \\
		\specialrule{0.2em}{0em}{0em}
		\lilanet{} VLS-128 \newline evaluation VLP-32C
		& $83.0\%$ & $30.8\%$ & $40.2\%$ & $4.8\%$ & $68.3\%$ & $22.7\%$ & $19.5\%$ & $63.8\%$ & $29.6\%$ & $\mathbf{46.3\%}$ & $68.1\%$ & $\mathbf{49.7\%}$ & $43.9\%$ \\
		\hline
		\pillarlikenet{} VLS-128 \newline evaluation VLP-32C
		& $\mathbf{86.2\%}$ & $\mathbf{50.4\%}$ & $\mathbf{75.9\%}$ & $\mathbf{36.5\%}$ & $\mathbf{80.8\%}$ & $\mathbf{34.0\%}$ & $\mathbf{28.4\%}$ & $\mathbf{69.5\%}$ & $\mathbf{35.4\%}$ & $35.5\%$ & $\mathbf{70.5\%}$ & $47.5\%$ & $\mathbf{54.2\%}$ \\
		\specialrule{0.5em}{0em}{0em}
		\lilanet{} VLP-32C \newline finetuned VLS-128 \newline evaluation VLS-128
		& $87.8\%$ & $\mathbf{58.0\%}$ & $46.0\%$ & $13.6\%$ & $77.1\%$ & $14.1\%$ & $25.0\%$ & $81.5\%$ & $39.0\%$ & $\mathbf{46.0\%}$ & $\mathbf{84.4\%}$ & $\mathbf{65.2\%}$ & $53.2\%$ \\
		\hline
		\pillarlikenet{} VLP-32C \newline finetuned VLS-128 \newline evaluation VLS-128
		& $\mathbf{88.5\%}$ & $\mathbf{58.0\%}$ & $\mathbf{64.9\%}$ & $\mathbf{42.5\%}$ & $\mathbf{87.3\%}$ & $\mathbf{49.5\%}$ & $\mathbf{36.9\%}$ & $\mathbf{84.4\%}$ & $\mathbf{45.6\%}$ & $45.1\%$ & $83.6\%$ & $63.1\%$ & $\mathbf{62.5\%}$ \\
		\specialrule{0.2em}{0em}{0em}
		\lilanet{} VLS-128 \newline finetuned VLP-32C \newline evaluation VLP-32C
		& $92.2\%$ & $66.3\%$ & $63.8\%$ & $29.1\%$ & $83.1\%$ & $45.6\%$ & $36.5\%$ & $\mathbf{80.3\%}$ & $41.1\%$ & $75.8\%$ & $\mathbf{80.1}\%$ & $\mathbf{63.8\%}$ & $63.1\%$ \\
		\hline
		\pillarlikenet{} VLS-128 \newline finetuned VLP-32C \newline evaluation VLP-32C
		& $\mathbf{92.6\%}$ & $\mathbf{66.6\%}$ & $\mathbf{77.6\%}$ & $\mathbf{53.1\%}$ & $\mathbf{86.8\%}$ & $\mathbf{58.8\%}$ & $\mathbf{39.0\%}$ & $78.9\%$ & $\mathbf{43.6\%}$ & $\mathbf{76.3\%}$ & $77.3\%$ & $62.3\%$ & $\mathbf{67.7\%}$ \\
		\hline
	\end{tabular}	
\end{table*}

\subsection{Same-Sensor Evaluation}
\label{sec:sameSensorEvaluation}

First, the networks are trained as described in \cref{sec:experiments},
where training, fine-tuning as well as evaluation is performed on
the same dataset, but on different subsets.
The second block of \cref{tab:overallResults} shows the results
of this same-sensor evaluation strategy.
The proposed \pillarlikenet{} clearly outperforms \lilanet{} on the
VLS-128 dataset and reaches slighly better results on the
VLS-32C dataset as well. This indicates that the voxel representation
outperforms the cylindrical 2D representation in terms
of output quality. However, compared to \lilanet{},
the inference time of \pillarlikenet{}
increases by a factor of 7 for the VLP-32C and a factor of 2 for the VLS-128.
This is mainly caused by the required voxelization and the larger dimensions
of the backbone CNN input pseudo-image.

Interestingly, both network architectures
achieve better results on the VLP-32C dataset than on the VLS-128 dataset.
This effect is mainly due to the smaller overall size of the VLS-128 dataset.
Also, the decrease in performance is larger for \lilanet{} than for \pillarlikenet{}.
This can be attributed to the higher resolution of the VLS-128, which directly influences
the object sizes within the cylindrical point cloud representation.
This indicates that the \pillarlikenet{} architecture is more suitable for transfer between sensors
than \lilanet{}.

\subsection{Cross-Sensor Evaluation}
\label{sec:crossSensorEvaluation}

Using the already trained networks of \cref{sec:sameSensorEvaluation},
the second evaluation stage is performed on the data of the opposite sensor
in order to evaluate cross-sensor portability.
The corresponding results are listed in the third block of \cref{tab:overallResults}.
\pillarlikenet{} clearly outperforms \lilanet{} by
more than 10 percentage points, confirming that the voxel representation
results in a far more portable architecture than the cylindrical projection.
At the same time
the mean IoU of \pillarlikenet{} drops by more than 16 percentage points
compared to the same-sensor evaluation results.
This drop might in part be due to the backbone CNN, which has
to handle strongly varying densities for the different sensor types and resolutions.
While the cross-sensor results are very promising, we note that there is still ample room for tuning
the voxel representation for portability.

\subsection{Cross-Sensor Finetuning}
\label{sec:crossSensorFinetuning}

As seen in \cref{sec:crossSensorEvaluation}, the direct application
of network models to different sensors leads to a significant drop in the output performance.
Therefore, we propose a data-driven adaptation step where the pre-trained model is fine-tuned
on the target sensor using manually annotated data.
Note that only a small amount of manually annotated data is required,
while the the full amount of autolabeled data of the target sensor is not used.
The results of this strategy are shown in the fourth block of \cref{tab:overallResults}.
After fine-tuning the network architecture on the target sensor,
\pillarlikenet{} still outperforms
\lilanet{}, which once more confirms the superior portability of the
\pillarlikenet{} architecture, allowing for an adaptation to
the target sensor type with only small amounts of additional data.

When compared to pure \autolabeling{} this training strategy
increases the mean IoU of \pillarlikenet{} by 2.4
percentage points on the VLS-128 dataset and by 3.8 percentage points
on the VLP-32C data. This shows that the presented adaptation process
can be used to successfully transfer network architectures across sensors
by applying only a small manually annotated dataset for fine-tuning instead
of using another sensor modality such as cameras to generate reference data.
It is conceivable that a fine-tuned \pillarlikenet{} can be used to extend the
\autolabeling{} concept of \cite{Piewak2018}, which originally relies on an additional sensor modality,
in order to automatically generate large-scale datasets for new sensor types of the same
modality (e.g. LiDAR to LiDAR \autolabeling{}).

\section{CONCLUSION}
\label{sec:conclusion}

In this paper, we present a CNN architecture
for the fine-grained semantic segmentation of LiDAR point clouds
based on a pillar-like voxel representation.
The proposed architecture is designed for portability
across different LiDAR sensor types to successfully handle
varying spatial resolution and scanning patterns.
We evaluate the network architecture against
a state-of-the-art semantic segmentation approach based on a
cylindrical projection of the LiDAR data \cite{Piewak2018}.
Our evaluation on manually annotated data across different sensors
shows that the proposed architecture is indeed highly portable,
yielding an improvement of 10 percentage points in IoU
when compared to the reference approach of \cite{Piewak2018}.
However, the employed voxel representation leads to an increase in
computational complexity, resulting in significantly longer inference times.

We further show that the presented architecture
can be fully transfered across different sensor types with minimal adaptation effort
by fine-tuning the pre-trained network on a small target sensor dataset.
This represents a significant advantage given
the fast-paced development of LiDAR hardware technology.
The results indicate that the proposed network architecture
can provide an efficient way for the automated generation of large-scale training data
for novel LiDAR sensor types without the need for a multi-modal sensor setup.
Hence, it might complement or even replace the multi-modal \autolabeling{} method of \cite{Piewak2018}.

Additional measures to further increase the cross-sensor portability
of the considered methods include the optimization of the voxel layout
and the backbone CNN architecture as well as classical data
augmentation methods.


\bibliographystyle{IEEEtran}

\end{document}